\documentclass[journal]{IEEEtran}

\ifCLASSINFOpdf
   \usepackage[pdftex]{graphicx}
   \graphicspath{{../pdf/}{../jpeg/}}
   \DeclareGraphicsExtensions{.pdf,.jpeg,.png}
\else
   \usepackage[dvips]{graphicx}
   \graphicspath{{../eps/}}
   \DeclareGraphicsExtensions{.eps}
\fi

\ifCLASSOPTIONcompsoc
  \usepackage[caption=false,font=sf,labelfont=sf,textfont=sf]{subfig}
\else
  \usepackage[caption=false,font=footnotesize]{subfig}
\fi

\usepackage[noadjust]{cite}
\usepackage{tikz}
\usepackage{csquotes}
\usepackage{amsmath}
\usepackage{framed}
\usepackage{color}
\usepackage{hyperref}

\usepackage{algorithm}
\usepackage{algpseudocode}
\usepackage{pifont}
\author{Caner Sahin\textsuperscript{1}, Rigas Kouskouridas\textsuperscript{2} and Tae-Kyun Kim\textsuperscript{1}
\thanks{\textsuperscript{1}Imperial Computer Vision and Learning Lab (ICVL), Imperial College London, \textsuperscript{2}Wirewax {\tt\small \{c.sahin14, tk.kim\}@imperial.ac.uk}, {\tt\small rkouskou@gmail.com}}%
}

\begin{document}

\title{A Learning-based Variable Size Part Extraction Architecture for 6D Object Pose Recovery in Depth}



\maketitle

\begin{abstract}
State-of-the-art techniques for 6D object pose recovery depend on occlusion-free point clouds to accurately register objects in 3D space. To deal with this shortcoming, we introduce a novel architecture called \textit{Iterative Hough Forest with Histogram of Control Points} that is capable of estimating the 6D pose of occluded and cluttered objects given a candidate 2D bounding box. Our \textit{Iterative Hough Forest (IHF)} is learnt using parts extracted only from the positive samples. These parts are represented with \textit{Histogram of Control Points (HoCP)}, a \enquote{scale-variant} implicit volumetric description, which we derive from recently introduced Implicit B-Splines (IBS). The rich discriminative information provided by the scale-variant HoCP features is leveraged during inference. An automatic variable size part extraction framework iteratively refines the object's initial pose that is roughly aligned due to the extraction of coarsest parts, the ones occupying the largest area in image pixels. The iterative refinement is accomplished based on finer (smaller) parts that are represented with more discriminative control point descriptors by using our \textit{Iterative Hough Forest}. Experiments conducted on a publicly available dataset report that our approach show better registration performance than the state-of-the-art methods.\\


\end{abstract}
\vspace{-1em}
\section{Introduction}
Object registration is an important task in computer vision that determines the position and the orientation of an object in camera-centered coordinates \cite{39}. An object of interest that was detected beforehand in a coarse 2D bounding box is fed into a registration system that can superimpose the desired translation and rotation of the object onto the raw camera image. By utilizing such a system, one can propose promising solutions for various problems related to scene understanding, augmented reality, control and navigation of robotics, \textit{etc}. Recent developments on visual depth sensors and their increasing ubiquity have allowed researchers to make use of the information acquired from these devices to facilitate challenging registration scenarios.\\
\indent Iterative Closest Point (ICP) algorithm \cite{1}, point-to-model based methods \cite{2, 3} and point-to-point techniques \cite{4, 5} demonstrate good registration results. However, the performance of these approaches is severely degraded in cases of heavy occlusion and clutter, and similar looking distractors. In order to address these challenges, several learning based methods formulate occlusion aware features \cite{6}, derive patch-based (local) descriptors \cite{15} or encode the contextual information of the objects with simple depth pixels \cite{8} and integrate them into random forests. Most particularly, iterative random forest algorithms such as Latent-Class Hough forest (LCHF) \cite{15} and iterative Multi-Output Random forest (iMORF) \cite{9} obtain state-of-the-art accuracy on pose estimation. On the other hand, these methods rely on scale-invariant features, while the exploitation of rich discriminative information inherently embedded into the scale-variability is one important point been overlooked.\\
\begin{figure*}[!t]
\captionsetup[subfigure]{labelformat=empty}
\centering
\includegraphics[height=2.6in]{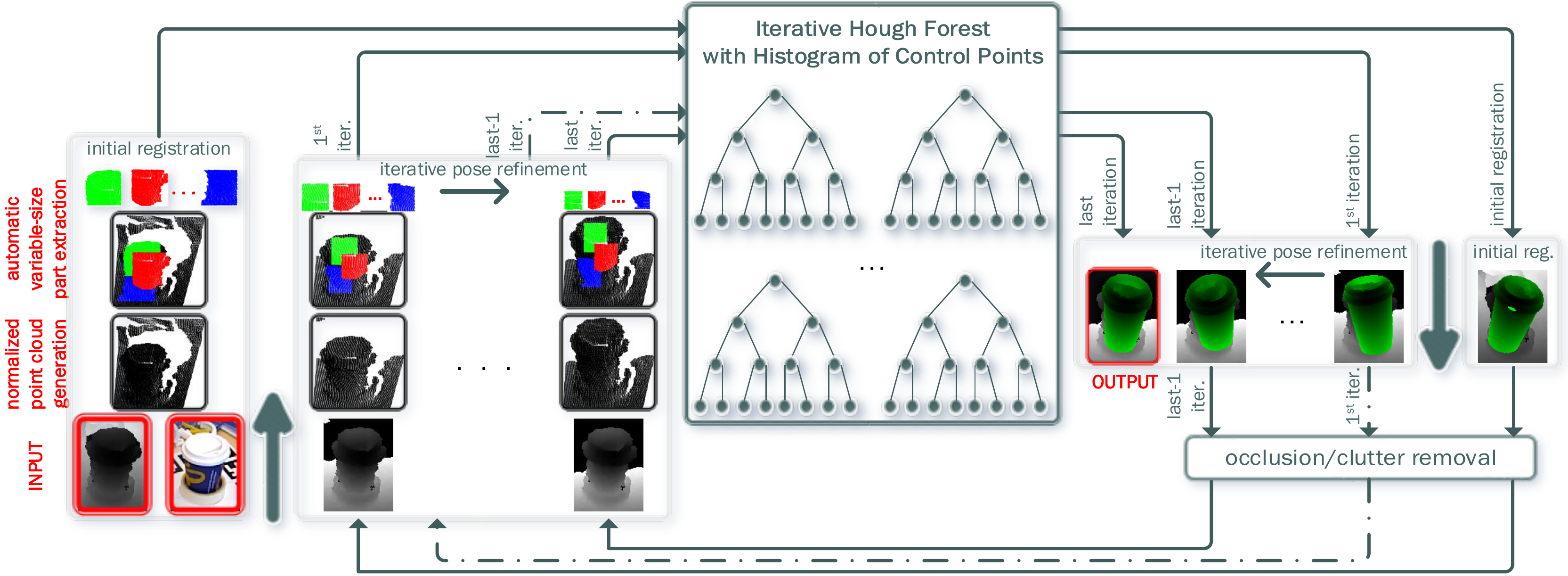}
\caption{Sample result of our architecture: the object of interest (lower-left corner) is first roughly aligned by extracting coarsest parts, the ones occupying the largest area in image pixels. This alignment is then iteratively refined based on finer (smaller) parts that are represented with more discriminative descriptors (The RGB image of the object of interest is for visualization purposes, the color-coded parts are centered on the same pixels).}
\vspace{-1em}
\label{fig1}
\end{figure*}
\indent Unlike the aforementioned learning-based methods, the ones presented by Novatnack \textit{et al.} \cite{10, 11} utilize the detailed information of the scale variation in order to register range images in a coarse-to-fine fashion. Although promising, they extract and match conventional salient 3D key points. However, real depth sensors have several imperfections such as missing depth values, noisy measurements, foreground fattening, \textit{etc}. As a result, salient feature points used in \cite{10} tend to be located on these deficient parts of the depth images, and hence, they are rather unstable \cite{12}. In such a scenario, 3D reconstruction methods that provide more reliable shape information can be utilized \cite{6}. Implicit B-Splines (IBS) \cite{7, 13} are techniques that can provide shape descriptors through their zero-sets and reconstruct surfaces. They are based on locally controlled functions that when combined with their control points produce a very rich part-based object representation.\\
\indent Our architecture is originated from these observations. We integrate the coarse-to-fine registration approach presented in \cite{10} into the random forest framework \cite{7, 9} using Histogram of Control Points (HoCP) features that we adapt from recently introduced IBSs \cite{13}. We train our forest only from positive samples and learn the detailed information of the scale-variability during training. We normalize every training point cloud into a unit cube and then generate a set of scale-space images, each of which is separated by a constant factor. The parts extracted from the images in this set are represented with the scale-variant HoCP features. During inference, the parts centered on the pixels that belong to the background and foreground clutters are removed iteratively using the most confident hypotheses and the test image is updated. Since this removal process decreases the standard deviation of the test point cloud, subsequent normalization applied to the updated test image increases the relative scale of the object (foreground pixels) in the unit cube. More discriminative control point descriptors are computed at higher scales and this ensures the refinement of the object pose. In our prior work \cite{14}, we have evaluated the registration performance of the proposed architecture by only using fixed-size parts. We extend the work engineering an automatic variable size part extraction framework in such a way that we can further exploit the discriminative information provided by the HoCP features. This framework first roughly aligns the object of interest by extracting coarsest parts, the ones occupying the largest area in image pixels, and then iteratively refines its alignment based on finer (smaller) parts that are represented with more discriminative control point descriptors. Note that we employ a compositional approach, \textit{i.e.}, we concurrently detect the object in the target region and estimate its pose by aligning the parts in order to increase robustness across clutter. Figure \ref{fig1} depicts a sample result of our architecture. To summarize, our main contributions are as follows:
\begin{itemize}
\item To the best of our knowledge this is the first time an implicit object representation, Implicit B-Spline, is adapted into a \enquote{scale-variant} part descriptor and is associated with the random forests.
\item We introduce a novel iterative algorithm for the Hough forests: it finds out an initial hypothesis and improves its confidence iteratively by extracting more discriminative \enquote{scale-variant} descriptors due to the elimination of the background/foreground clutter.
\item We engineer an automatic variable size part extraction framework for the random forests: it first roughly aligns the object of interest by extracting coarsest parts and iteratively improves its confidence based on finer (smaller) ones.
\end{itemize}
The rest of the paper is organized as follows: In Sect. \ref{Related Work}, we present a review on the object registration. Section \ref{approach} demonstrates the computation procedure of the HoCP features as a scale-variant part representation, their combination with the Iterative Hough Forest (IHF), and the registration process. Experimental results are provided in Sect. \ref{experiments} and finally, the paper is concluded in Sect. \ref{conclusion} with several remarks, and discussions.
\vspace{-1em}
\section{Related Work}
\label{Related Work}
A large number of studies have been proposed for the object registration, ranging from the point-wise correspondence based methods to the learning based approaches. Iterative Closest Point (ICP) algorithm, originally presented in \cite{1}, requires a good initialization in order not to be trapped in a local minimum during fine tuning. This issue is addressed in \cite{21} providing globally optimal registration by the integration of a global branch-and-bound (BnB) optimization into the local ICP. The point-wise correspondence problem is converted into a point-to-model registration in \cite{2}. The object model is represented with implicit polynomials (IP) and the distance between the test point set and the object model is minimized via the Levenberg-Marquardt algorithm (LMA). Zheng \textit{et al.} \cite{22} propose a 6 DoF pose estimation technique utilizing 3D IPs on ultrasound images. In the off-line phase, object model is represented with 3D IPs and by utilizing its gradient flow, 2D ultrasound image is registered in the on-line process. In \cite{23}, a coarse-to-fine fast IP-driven registration method is presented. A rough pose estimation is quickly acquired with a coarse IP model (low degree curve fitting) and finer models refine the parameters of this rough estimation (high degree curve fitting). Hinterstoisser et al. \cite{36} extract holistic templates from 3D models of the objects and match to the scene at test time. These studies have demonstrated good registration results on the target point sets that are occlusion-free and/or are subjected to the artificial Gaussian noises and outliers.\\
\indent Unlike the abovementioned methods, more realistic registration scenarios have been addressed by the point-to-point techniques that build point-pair features for sparse representations of the test and the model point sets \cite{30, 31, 38}. Rusu \textit{et al.} align two noisy point clouds of real scenes by finding correct point-to-point correspondences between the Point Feature Histograms (PFH) and feed this alignment to an ICP algorithm for fine tuning \cite{29}. The cluttered and partially occluded objects' poses are hypothesized by accumulating the votes of the matching features in \cite{30}. Choi \textit{et al.} \cite{31} propose point-pair features for both RGB and depth channels that are conducted in a voting scheme to hypothesize the rotation and translation parameters of the objects in the cluttered scenes. The features proposed in \cite{5} make use of the visibility context of the scene to tackle the registration. Despite achieving good registration results, these techniques underperform when the scenes are under heavy occlusion and clutter, and the objects' geometry are indistinguishable from background.\\
\indent Learning-based methods have good generalization across severe occlusion and clutter \cite{6, 8, 32, 33}. The method presented in \cite{6} formulates the recognition problem globally and derives occlusion aware features. A set of principal curvature ratios are computed for all pixels in depth images to extract the edgelets. In \cite{8}, the contextual information of the objects is encoded with simple depth and RGB pixels. This technique improves the confidence of a pose hypothesis using a Ransac-like algorithm. Cabrera \textit{et al.} \cite{33} back project the parts inside the initially found coarse bounding box to the image and pass down the forest again. The parts with the lowest contributions are penalized in such a way that finer registration is produced.\\
\indent The state-of-the-art accuracy on registration is acquired by the iterative random forest algorithms. The part-based strategy, Latent-Class Hough Forest \cite{15}, refines the initially hypothesized object pose by iteratively updating the object class distributions in the leaf nodes during testing. Iterative Multi Output Random forest (iMORF) \cite{9} jointly predicts the head pose, the facial expression and the landmark positions. The relations between these tasks are modelled so that their performances are iteratively improved with the extraction of more informative features. The ideas, iterative pose refinement during testing and iterative extraction of more discriminative features, form a basis for our Iterative Hough Forest (IHF) architecture: during training, we encode discriminative shape information of the HoCP features into the forest. Despite that the skeleton of our training procedure is similar to the methodology in \cite{15}, our forest learns the discriminative shape information that will be iteratively exploited at test time. In the course of inference, unlike \cite{15}, we update the test image itself and the hypotheses confidence by a noise removal process that allows us to extract more informative features from the test images. Whilst these approaches \cite{9, 15} rely on the scale-invariant features to improve the confidence of a pose hypothesis, Novatnack \textit{et al.} \cite{10, 11} introduce a framework that registers the range images in a coarse-to-fine fashion by utilizing the detailed information provided by the scale variation. The shape descriptors with the coarsest scale are matched initially and a rough alignment is achieved since fewer features are extracted in coarser scales. The descriptor matching at higher scales produces improved predictions of the pose. Inspired by \cite{10}, we design a \enquote{scale-variant} implicit volumetric part description, \enquote{Histogram of Control Points (HoCP)}, and associate it with the random forest framework.\\
\indent Selecting the part size is important since larger parts tend to match the disadvantages of a holistic template while smaller ones are prone to noise \cite{15}. In heavily occluded and cluttered scenes, relatively smaller parts perform well whilst the larger ones are more convenient in occlusion/clutter-free scenarios. Discriminative information encoded into small sized parts might not be fully exploited by larger parts, most particularly when the object representation is scale-variant. Hence, this application-specific/task-dependent part size selection degrades the generalization and it is one of the remaining challenges that should be addressed, apart from occlusion, clutter and/or similar looking distractors. Beyond object pose estimation \cite{14, 15}, there are several part-based solutions proposed for different tasks such as human pose recognition \cite{17, 18}, 3D face analysis \cite{19}, or hand pose estimation \cite{20} to name a few. They experience different part sizes and select the one that performs best, however, none of these solutions investigate how extracting variable size parts can be utilized in a single framework. In this study, we investigate the effect of this \textit{size variation} and show that the simultaneous utilization of the parts of varying size can improve 6D object pose estimation, especially in heavily occluded and cluttered depth maps, supplying a rich source of discriminative information.
\section{Our Registration Approach}
\label{approach}
In this section we detail our registration approach by firstly describing the computation procedure of the HoCP features as a scale-variant part representation. We then present how to encode the discriminative information of these scale-variant features into the forest. Finally, we demonstrate how to exploit the learnt shape information in a coarse-to-fine fashion to refine the pose hypotheses.
\subsection{Scale-variant Part Representation: Histogram of Control Points}
\label{varpart}
Given a positive depth image we initially normalize it into a unit cube and then new point clouds at different scales are sampled as follows:
\begin{equation}
  \{ {\mathbf{X}_N \}}_i= \frac{\mathbf{X}_{n \times 3} - \bar{\mathbf{X}}_{n \times 3}}{s_i * \mathbf{\alpha}} + 0.5, \quad i = 0, 1, 2, ..., m
  \label{eq1}
\end{equation}
with
\begin{equation}
\small
\mathbf{\alpha} = \max \left  \{
				  \begin{tabular}{ccc}
				  max($X$)-min($X$) \\
				  max($Y$)-min($Y$) \\
				  max($Z$)-min($Z$) 
				  \end{tabular}
					  \right \}, h_i = \max(Z_{N_i})-\min(Z_{N_i})
					  \label{eq2}
\end{equation}
where $\textbf{X} = [X, Y, Z]$ is the world coordinate vector of the original foreground point cloud, $\bar{\mathbf{X}}$ is the mean of $\textbf{X}$, $\textbf{X}_N = [X_N,  Y_N,  Z_N]$ is the normalized foreground pixels, $m$ is the number of the scales, $\alpha$ is the scale factor and $h$ is the scale. The constant $s_i$ takes real numbers to generate the point clouds at different scales, starting from $s_0 = 1$ that corresponds to the initial normalization. A training image and its samples at different scales are shown in Fig. \ref{fig2} (a).\\
\indent Once we generate a set of scale-space images, we represent these point clouds first globally with the control points of Implicit B-Splines (IBS). IBS is defined through the combination of B-Spline tensor products:
\begin{figure*}[!t]
\captionsetup[subfigure]{labelformat=empty}
\centering
\includegraphics[height=3in]{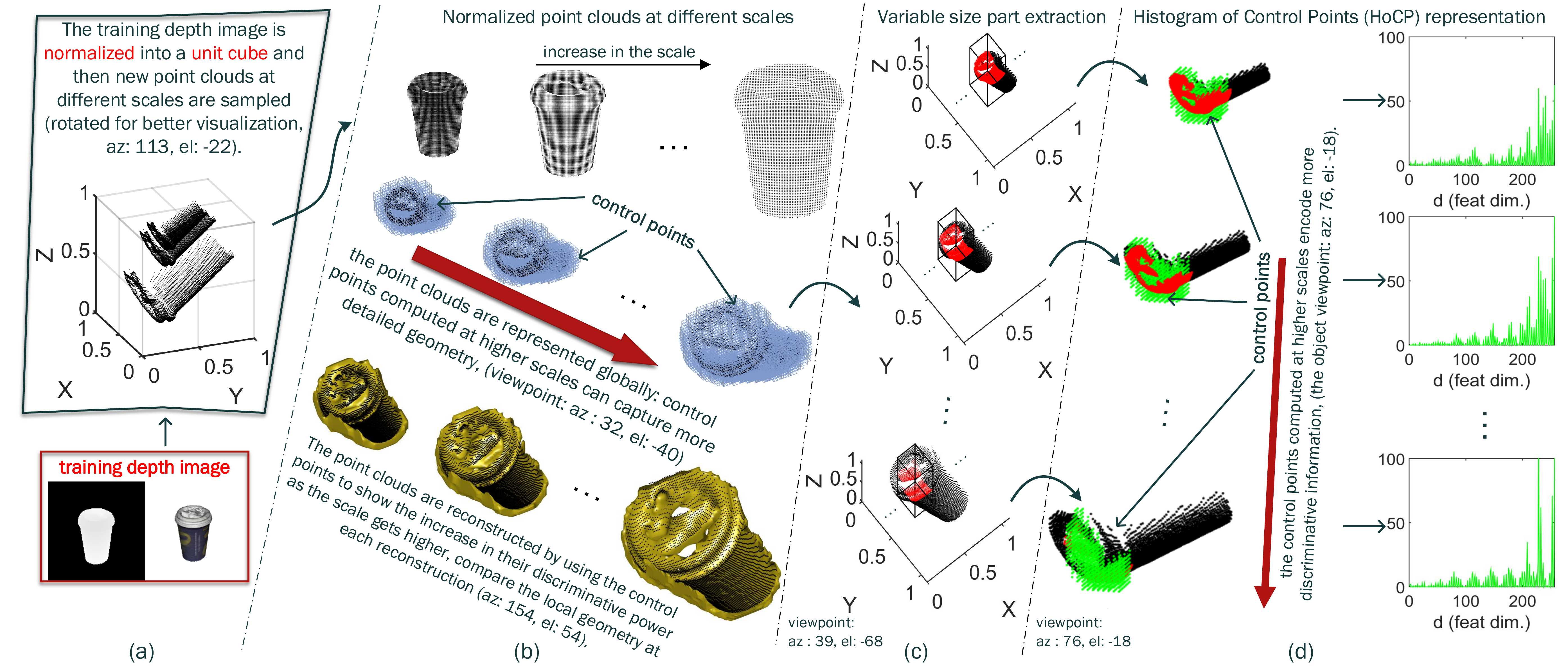}
\caption{Computation procedure of the HoCP features as a scale-variant part representation: (a) initial normalization ($s_0 = 1$) of the training depth image is the outmost point cloud and the inner ones are sampled by different $s_i$ values. (b) global representation of the scale-space images. (c) extraction process of the variable size parts (centers of these parts are the same). (d) HoCP representation of the parts extracted in (c).}
\label{fig2}
\end{figure*}
\vspace{-0.5em}
\begin{equation}
f(\mathbf{x}) = \sum_{i,j,k = 1} ^N n_{i,j,k} B_i(x) B_j(y) B_k(z)
  \label{eq3}
\end{equation}
where $\{n_{i,j,k}\}$ are the coefficients defining a control lattice of size $N \times N \times N$ and $B_i(x)$, $B_j(y)$, $B_k(z)$ are the spline basis functions. This definition can be reformulated as the following inner product:
\vspace{-0.5em}
\begin{equation}
f(\mathbf{x}) = \mathbf{n}^T \mathbf{e(x)} = \mathbf{e(x)}^T \mathbf{n}
  \label{eq4}
\end{equation}
where the coefficient vector $\mathbf{n}$ includes the control values $\{n_{i,j,k}\}$ and the basis vector $\mathbf{e(x)}$ depends on the given data point since it sorts the spline basis function products $\{B_i(x)B_j(y)B_k(z)\}$. The basis vectors in Eq. \ref{eq4} are computed for the whole point cloud and the coefficient vector $\mathbf{n}$ is calculated based on the 3L algorithm \cite{37}. Rouhani et al. \cite{13} construct the spline basis functions $B_i(x)$, $B_j(y)$, $B_k(z)$ through the following blending functions:
\begin{equation}
\begin{split}
b_0(u) = (1-u^3)/6, \quad b_1(u) = (3u^3-6u^2+4)/6\\
b_2(u) = (-3u^3 + 3u^2 + 3u + 1)/6, \quad b_3(u) = u^3/6
\end{split}
  \label{eq5}
\end{equation}
and reformulate Eq. \ref{eq3} in order to determine the control point vector $\mathbf{n}$ of the point clouds that are normalized into the unit cube $[0 \quad 1]^3$:
\vspace{-0.5em}
\begin{equation}
f(\mathbf{x}) = \sum_{l,m,p = 0} ^3 n_{i+l,j+m,k+p} b_l(u) b_m(v) b_p(w)
  \label{eq6}
\end{equation}
where
\begin{gather*} 
i = \lceil x/ \Delta \rceil, \quad j= \lceil y/ \Delta \rceil, \quad k= \lceil z/ \Delta \rceil\\
u= \frac{x}{\Delta} - \lfloor \frac{x}{\Delta} \rfloor, \quad v = \frac{y}{\Delta} - \lfloor \frac{y}{\Delta} \rfloor, \quad w= \frac{z}{\Delta} - \lfloor \frac{z}{\Delta} \rfloor\\
\Delta = 1/(N-3).
\end{gather*}
Thus, the unit cube is split into an $ N \times N \times N $ voxel grid where $N$ is the IBS resolution. Each control point in $\mathbf{n}$ is defined with an index-weight pair: the index number indicates the vertex of this grid at which the related control point is located. The weight informs the descriptor significance about the control of the geometry to be represented. The scale-space images in Fig. \ref{fig2} (a) are globally represented in Fig. \ref{fig2} (b) with the control point descriptors. We use all control points to represent the structures, but one can sort these descriptors based on their weights and utilize the ones higher than a threshold. In Fig. \ref{fig2} (b), the point clouds are lastly reconstructed by using the control points to show the increase in their discriminative power as the scale gets higher. Note that IBS resolution $N$ determines the complexity of the representation (level of detail) in a unit cube, whilst the scale indicates the relative size of the object with respect to the unit cube dimensions. In our architecture, despite sampling point clouds at different scales $h_i$, (\textit{e.g.} $i = 0, 1, ..., 8$), we fix the complexity of the representation, (\textit{e.g.} $N = 50$).\\
\indent IBS is the combination of the locally controlled functions and allows one to propose effective part-based solutions for object registration. We benefit from such a property and partition the globally represented scale-space depth maps into parts. We express the part size $g$ in image pixels which also depicts the ratio between the sizes of the extracted part and the bounding box of the global point cloud. In our prior work \cite{14}, we have extracted and represented the parts that have the same size, that is, the parts growing around every individual pixel at each scale occupy the same area in image pixels. We now extend the work extracting the parts that are different in size. A 3D bounding box defined in metric coordinates is traversed in the unit cube of each scale-space image and the parts are extracted around non-zero pixels. The total number of the data points in this 3D bounding box varies for the point clouds at different scales, and consequently, the size of the extracted parts differs. Figure \ref{fig2} (c) shows an example of variable size part extraction process in which the red parts are grown around the same data point. Note how the part size decreases when the scale of the normalized object point cloud gets higher, since less number of data points are deployed in the 3D bounding box. Each part has its own implicit volumetric representation, formed by the closest control points to the part center, the ones lying inside the 3D bounding box along depth direction. Such a part description characterizes the locality in a cascaded fashion, growing regions with different characteristics around a point. We encode this information into histograms in spherical coordinates. Each of the part centers is coincided with the center of a sphere. The control points of the part are described by the log of the radius $t_r$, the cosine of the inclination $t_{\theta}$ and the azimuth $t_{\phi}$. Then, the sphere is divided into the bins and the relation between the bin numbers $\nu_r, \nu_{\theta}, \nu_{\phi}$ and the histogram coordinates $t_r, t_{\theta}, t_{\phi}$ is given as follows \cite{21}:
\vspace{-0.5em}
\begin{equation}
\begin{split}
       t_r &= \frac{\nu_r}{log(\frac{r_{max}}{r_{min}})} log(\frac{r}{r_{min}}) \\
t_{\theta} &= \nu_{\theta} \frac{z}{r} \\
  t_{\phi} &= \frac{\nu_{\phi}}{2 \pi} tan^{-1}(\frac{y}{x})  \\
\end{split}
\end{equation}
where $r_{min}$ and $r_{max}$ are the radii of the nested spheres with the minimum and the maximum volumes, $x, y, z$ are the Cartesian coordinates of each descriptor with radius $r$. $r_{max}$ equals to the distance between the patch center and the farthest descriptor of the related patch. The numbers of the control points in each bin are counted and stored in a $d = \nu_r*\nu_{\theta}*\nu_{\phi}$ dimensional feature vector $\mathbf{f}$. Figure \ref{fig2} (d) illustrates the HoCP representations of the parts extracted in Fig. \ref{fig2} (c). Note that the control points computed at higher scales capture more detailed part geometry.
\subsection{The Combination of HoCP and Iterative Hough Forest}
The proposed IHF is the combination of randomized binary decision trees. It is trained only on foreground synthetically rendered depth images of the object of interest. We generate a set of scale-space images from each training point cloud and sample a set of parts $\{ P_i \}$ as explained in subsection \ref{varpart} and annotate those as follows:
\begin{equation}
\mathcal{P} = \{ P_i \} = \{ (\mathbf{c}_i, \Delta \mathbf{x}_i, \mathbf{\theta}_i, \mathbf{f}_i, D_i)\}
\label{eq18}
\end{equation}
where $\mathbf{c}_i = (c_{x_i}, c_{y_i})$ is the part center in pixels, $ \Delta \mathbf{x}_i = (\Delta x_i, \Delta y_i, \Delta z_i)$ is the 3D offset between the centers of the part and the object, $\theta_i = (\theta_{r_i}, \theta_{p_i}, \theta_{y_i}) $ is the rotation parameters of the point cloud from which the part $P_i$ is extracted and $D_i$ is the depth map of the part.\\
\indent Each tree is constructed by using a subset $\mathcal{S}$ of the annotated training parts $\mathcal{S} \subset \mathcal{P}$. We randomly select a template patch $T$ from $\mathcal{S}$ and assign it to the root node. We measure the similarity between $T$ and each patch $S_i$ in $\mathcal{S}$ as follows:
\begin{itemize}
\item \textbf{Depth check:} The depth values of the descriptors $S_i^\mathbf{n}$ and $T^\mathbf{n}$ that represent the parts $S_i$ and $T$ are checked, and the spatially inconsistent ones in $S_i^\mathbf{n}$ are removed as in \cite{7}, generating $\Omega$ that includes the spatially consistent descriptors of the patch $S_i$.
\item \textbf{Similarity measure:} Using $\Omega$, the feature vector $\mathbf{f}_{\Omega}$ is generated and the $\mathcal{L}_2$ norm between this vector and $\mathbf{f}_T$ is measured:
\begin{equation}
\mathcal{F}(S_i, T) = {\parallel \mathbf{f}_{\Omega} - \mathbf{f}_T \parallel}_2
\end{equation}
\item \textbf{Similarity score comparison:} Each patch is passed either to the left or the right child nodes according to the split function that compares the score of the similarity measure $\mathcal{F}(S_i, T)$ and a randomly chosen threshold $\tau$.
\end{itemize}
\indent The main reason why we apply a depth check to the patches is to remove the structural perturbations, due to occlusion and clutter \cite{7}. These perturbations most likely occur on patches extracted along depth discontinuities such as the contours of the objects. They force a test patch to diverge (occluded/cluttered) from its positive correspondence by changing its representation, $r_{max}$ of the sphere, and the histogram coordinates consequently.\\
\indent A group of candidate split functions are produced at each node by using a set of randomly assigned patches $\{ T_i \}$ and thresholds $\{ \tau_i \}$. The one that best optimize the offset and pose regression entropy \cite{22} is selected as the split function. Each tree is grown by repeating this process recursively until the forest termination criteria are satisfied. When the termination conditions are met, the leaf nodes are formed and they store votes for both the object center $\Delta \mathbf{x} = (\Delta x, \Delta y, \Delta z)$ and the object rotation $\theta = (\theta_r, \theta_p, \theta_y) $.\\
\indent Depending on the part extraction approach, all parts in $\mathcal{P}$ (see Eq. \ref{eq18}) can either be of the same size or of the variable size. From now on, we will refer to the forests trained on variable size parts as the \textit{IHF-variable size}, and to the ones learnt by using fixed size parts as the \textit{IHF-fixed size}.
\subsection{6D Object Pose Estimation}
\begin{figure}[!t]
\captionsetup[subfigure]{labelformat=empty}
\centering
\includegraphics[height=2.9in]{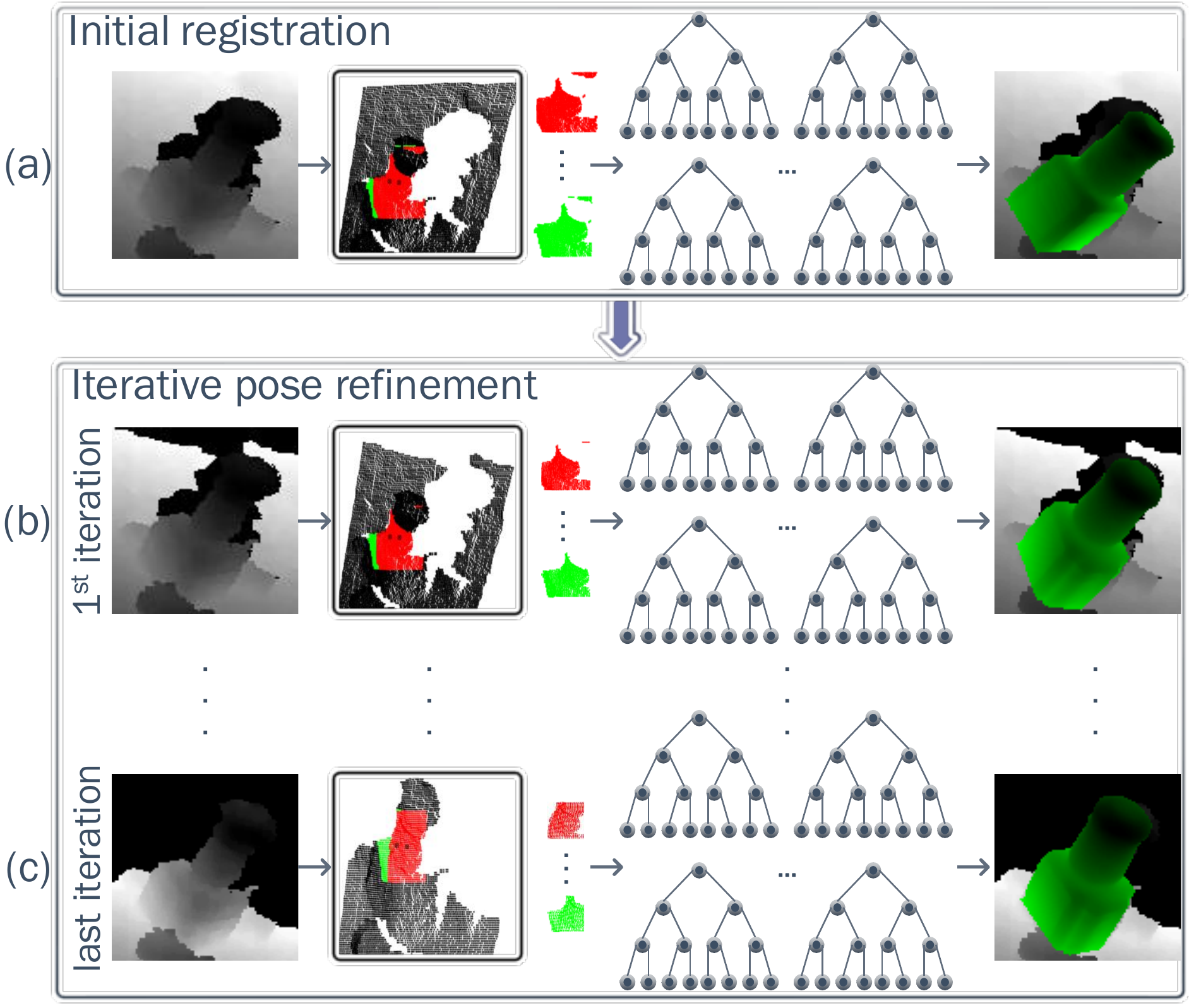}
\caption{The object of interest is first roughly aligned by extracting the coarsest parts and this alignment is then iteratively refined based on finer (smaller) ones.}
\vspace{-1.5em}
\label{fig3}
\end{figure}

Once we encode the discriminative information of the scale-variant HoCP features into the forest, we next demonstrate 6D pose estimation of objects considering that the learnt forest is IHF-variable size.\\
\indent The proposed architecture registers objects in two steps: the \textit{initial registration} and the \textit{iterative pose refinement}. The \textit{initial registration} roughly aligns the test object and this alignment is further improved by the \textit{iterative pose refinement}.\\
\indent Consider an object that was detected by a coarse bounding box, $I_b$, as shown in the leftmost image of Fig. \ref{fig3} (a). At an iteration instant $k$, the following quantities are defined:
\begin{itemize}
\item $\Delta \mathbf{x}^{0:k} = \{ \Delta \mathbf{x}^0, \Delta \mathbf{x}^1, ..., \Delta \mathbf{x}^k \} = \{ \Delta \mathbf{x}^0, \Delta \mathbf{x}^{1:k} \}$: the history of the object position predictions.
\item $\theta^{0:k} = \{ \theta^0, \theta^1, ..., \theta^k \} = \{ \theta^0, \theta^{1:k} \}$: the history of the object rotation estimations.
\item $V^{1:k} = \{ v^1, v^2, ..., v^k \}$ : the history of the inputs (noise removals) applied to the test image.
\item $ \mathcal{M}^{0:k} = \{ \mathcal{M}^0, \mathcal{M}^1, ..., \mathcal{M}^k \} = \{ \mathcal{M}^0, \mathcal{M}^{1:k} \}$: the history of the set of the feature vectors where $\mathcal{M}^k = \{ \mathbf{f}_i \}$.
\item $h^k$: the object scale (the scale of the foreground pixels) in the unit cube at iteration $k$ (see Eq. \ref{eq2}).
\item $g^k$: the size of the parts extracted at iteration $k$.
\end{itemize}
We formulate the \textit{initial registration} as follows:
\begin{equation}
( \Delta \mathbf{x}^0, \theta^0 ) = \arg \max_{\Delta \mathbf{x}^0, \theta^0}  p(\Delta \mathbf{x}^0, \theta^0 \vert I_b, \mathcal{M}^0, h^0, g^0).
\label{eq8}
\end{equation}
We find the best parameters that maximize the joint posterior density of the initial object position $\Delta \mathbf{x}^0$ and the initial object rotation $\theta^0$. The initial registration process is illustrated in Fig. \ref{fig3} (a). The test image is firstly normalized into a unit cube. Unlike training, this is a \enquote{single} scale normalization that corresponds to $s_0 = 1$ (see Eq. \ref{eq1}). The patches extracted from the globally represented point cloud are described with the HoCP features and are passed down all the trees. At this stage, we extract the coarsest patches from the test image, \textit{i.e.}, the ones occupying the largest area in image pixels. We determine the effect that all patches have on the object pose by accumulating the votes stored in the leaf nodes as in \cite{7} and approximate the initial registration given in Eq. \ref{eq8}. Once the initial hypothesis $ \mathbf{x}^0 = (\Delta \mathbf{x}^0, \theta^0)$ is obtained, the set of pixels that belong to the background/foreground clutter $\{ p_j \}$ are removed from $I_b$ according to the following criterion:
\begin{equation}
    v^k=
    \begin{cases}
      I_b(p_j) = \mathcal{D}_{I_b}(p_j), & \gamma \psi_1 < \mathcal{D}_{I_b} (p_j^k) < \beta \psi_2 \\
      I_b(p_j) = 0, & otherwise 
    \end{cases}
    \label{eq10}
  \end{equation}
  with
  \begin{equation}
  \gamma = \min (\mathcal{D}_H^k), \quad \beta = \max (\mathcal{D}_H^k)
  \end{equation}
where $\mathcal{D}_H^k$ and $\mathcal{D}_{I_b}$ are the depth maps of the hypothesis $H$ at iteration $k$, and of the $I_b$, $\psi_1$ and $\psi_2$ are the scaling coefficients. The efficacy of $v^k$ is illustrated in Fig. \ref{fig3}. In the rightmost image of Fig. \ref{fig3} (a), the test image and the initial hypothesis are superimposed. This hypothesis is exploited and the test image is updated by $v^1$ as in Eq. \ref{eq10}. The updated image is shown in Fig. \ref{fig3} (b) and assigned as input for the $1^{st}$ iteration. It is normalized and represented globally. The object \enquote{scale} ($h^1$) in the unit cube is relatively increased (compare with the initial registration) and more discriminative control point descriptors $\mathbf{n}$ are computed. This is mainly because of the fact that the standard deviation of the input image decreased since we removed foreground/background clutter. As a follow up step, we traverse the 3D bounding box in the unit cube during part extraction, while the increase in the normalized object scale gives rise to extract patches whose size are smaller (finer) than the ones extracted during the initial registration. The resulted hypothesis of the $1^{st}$ iteration is shown on the right. The extraction of finer parts represented with more discriminative control point descriptors along with the noise removal process result in more accurate and confident hypothesis. This pose refinement process is iteratively performed until the maximum iteration is reached (see Fig. \ref{fig3} (c)):
\begin{equation}
\small
( \Delta \mathbf{x}^k, \theta^k ) = \arg \max_{\Delta \mathbf{x}^k, \theta^k}  p(\Delta \mathbf{x}^k, \theta^k \mid \mathcal{M}^{1:k}, V^{1:k}, \mathbf{x}^0, h^k, g^k)
\label{Eq13}
\end{equation}
We approximate the registration hypothesis at each iteration by using the stored information in the leaf nodes as we do in the initial registration. If we would demonstrate the 6D object pose estimation considering that the learnt forest is the IHF-fixed size, the only difference in the formulation would be the part extraction viewpoint. Instead of traversing 3D bounding box in the unit cube, we would extract the parts with a predefined size in pixels, and at an iteration instant $k$, $g^k$ would remain the same as $g^0$. In the next section, we will compare the registration performances of the forests that are separately trained on fixed and variable size parts.
\section{Experimental Results}
\label{experiments}
There are several publicly available datasets \cite{15}, \cite{36} to test the performances of the object registration methods. For each object type in these datasets, a set of RGB-D test images are provided with ground truth object pose parameters. We have analysed these images and have found that \enquote{Coffee Cup}, \enquote{Camera} and \enquote{Shampoo} (included in the ICVL dataset \cite{15}) are some of the best demonstrable objects to test and to compare our registration architecture with the state-of-the-art methods since they are located in highly occluded and cluttered scenes. We further process the test images of these objects to generate a new test dataset according to the following criteria:
\begin{figure}[!t]
\captionsetup[subfigure]{labelformat=empty}
\centering
\includegraphics[height=2.2in]{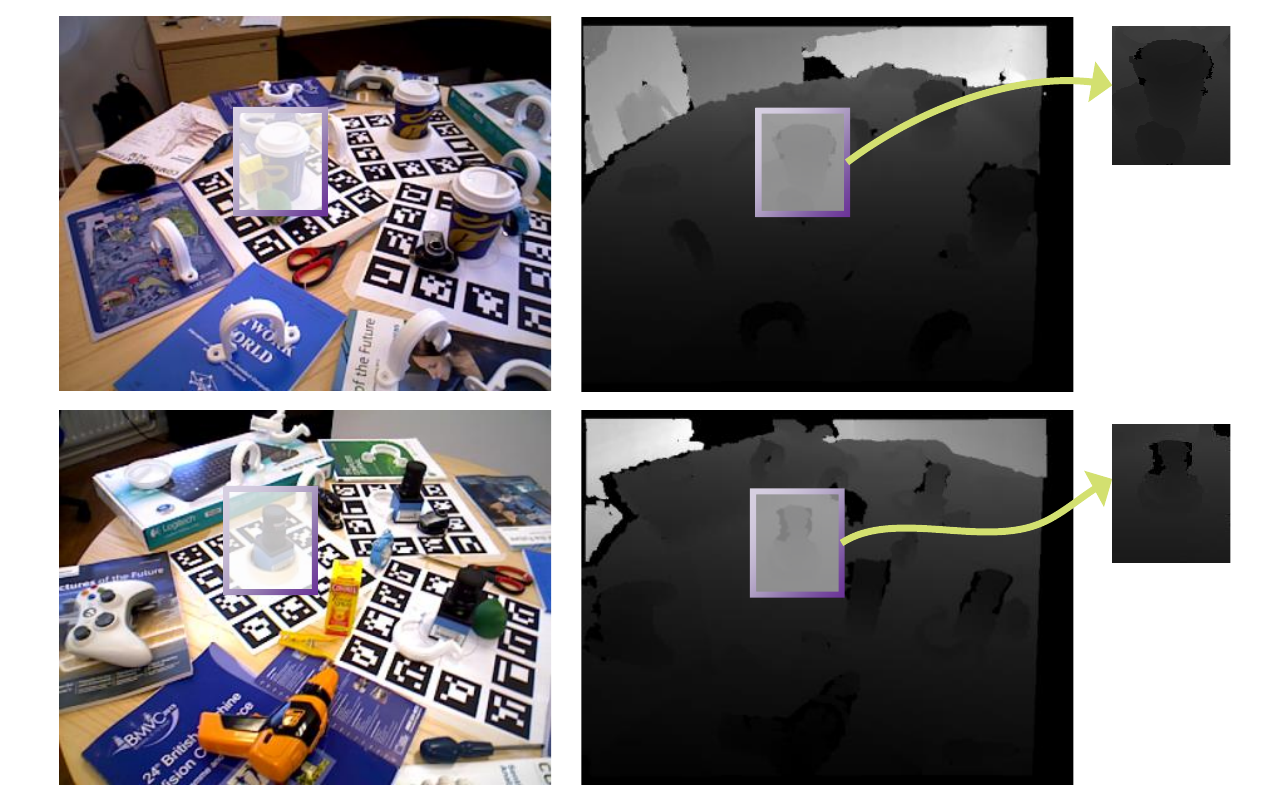}
\vspace*{-4mm}
\caption{Dataset generation: the input of the proposed framework is the depth image of the coarsely detected objects (RGB correspondence is for better visualization).}
\label{fig4}
\end{figure}
\begin{itemize}
\item Since the HoCP features are scale-variant, the depth values of the training and the test images should be close to each other up to a certain degree. In this study, we train the forests at a single depth value, $f_d$ mm, and test with the images at the range of $[f_d \mp 50]$ mm.
\item The test object instances located at the range of $[f_d \mp 50]$ mm are assumed as detected by coarse bounding boxes (see Fig. \ref{fig4}). The image regions included in these bounding boxes are cropped and the new test dataset is generated.
\end{itemize}
The generated dataset includes $276$ \enquote{Coffee Cup}, $360$ \enquote{Camera} and $200$ \enquote{Shampoo} images each of which is at the range of $[750 \mp 50]$ mm since we train the forests used in all experiments with the positive samples at $f_d = 750$ mm depth. The maximum depth is $25$ and the number of the maximum samples at each leaf node is $15$ for each tree. Every forest is the ensemble of $3$ trees with these termination criteria.\\
\begin{figure}[!t]
\captionsetup[subfigure]{labelformat=empty}
\centering
\includegraphics[height=5.5in]{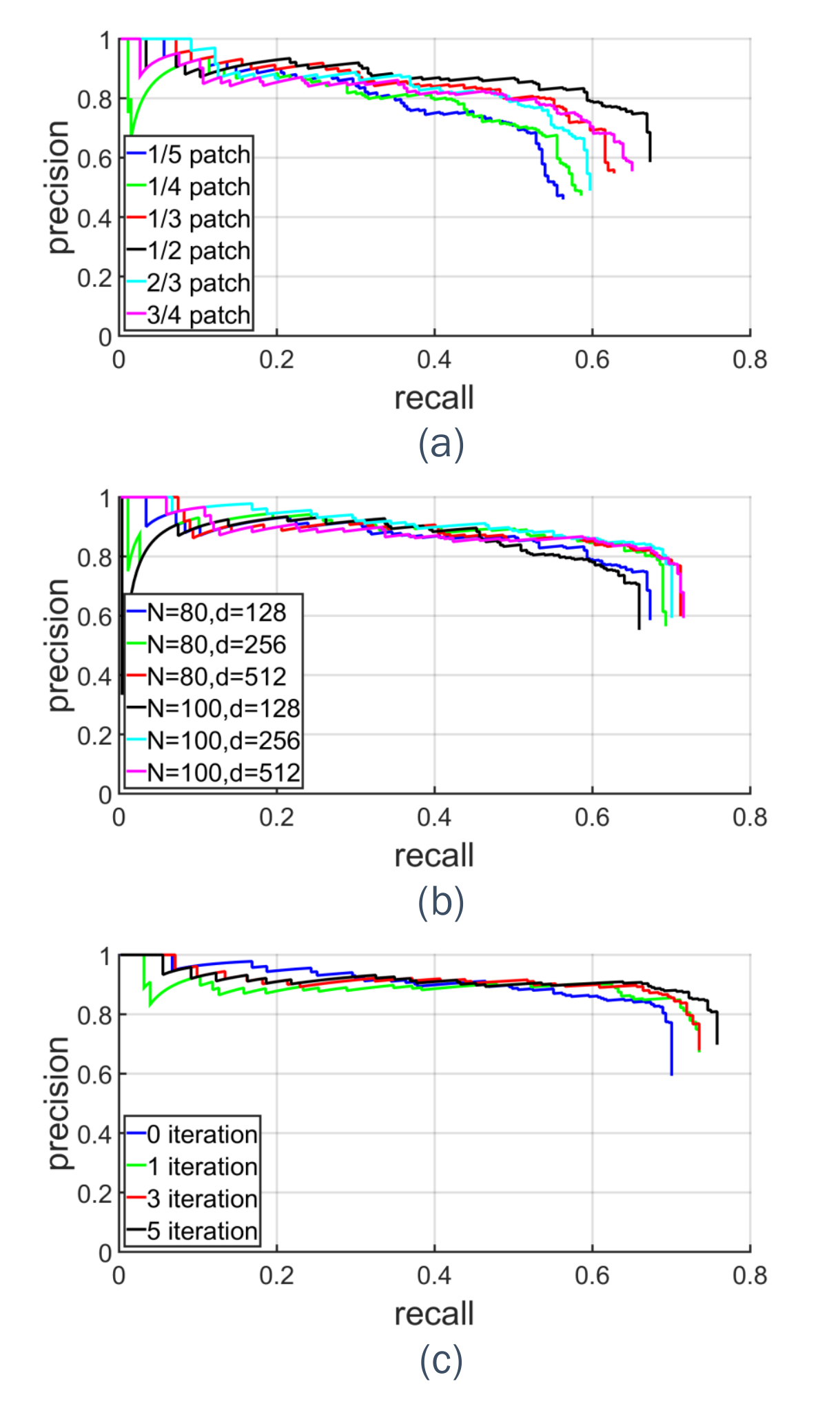}
\vspace*{-4mm}
\caption{Precision-Recall curves for parameter optimization: (a) compares the performances of the forests with different patch sizes. (b) illustrates the registration performances for different IBS resolutions $N$ and feature dimensions $d$. (c) shows the effect of the iteration number. For the corresponding F1 scores see Table \ref{table_params}.}
\label{fig5}
\end{figure}
\indent Our experiments are two folds: \textit{parameter optimization} and \textit{comparative study}. The architecture parameters have an important impact upon the registration and include the size of the parts extracted during the initial registration $g^0$, the IBS resolution $N$, the HoCP feature dimension $d$ (the number of the bins or quantization parameter) and the iteration number. Once the best parameters are acquired, we compare the performance of our architecture with the state-of-the-art methods in the comparative study.\\
\indent Both experiments use the metric proposed in \cite{36} to determine whether a registration hypothesis is correct. This metric outputs a score $\omega$ that calculates the distance between the ground truth and estimated poses of the test object. The registration hypothesis that ensures the following inequality is considered as correct:
\begin{equation}
\omega \leq z_{\omega} \Phi
\end{equation}
where $\Phi$ is the diameter of the 3D model of the test object and $z_{\omega}$ is a constant that determines the coarseness of an hypothesis that is assigned as correct. We set $z_{\omega}$ to $0.08$ when we refine the parameters of the proposed architecture and the effect of various $z_{\omega}$ values is separately examined in the comparative study.
\begin{table}[!t]
\caption{F1 scores of the initial registrations determined for different part sizes (g), IBS resolution (N) \& feature dimension (d) and number of iteration}
\centering
\begin{tabular}[t]{|c|c||c|c|}
\hline
$\textbf{Part}$&$ \textbf{F1}$   & $\textbf{N}$ \& $\textbf{d}$ &$ \textbf{F1}$\\
$\textbf{Size, g}$ &$ \textbf{Score}$& $ $                          &$ \textbf{Score}$\\
\hline
$\frac{1}{5}$         &0.5966           &80 \& 128                         &0.7068\\
\hline
$\frac{1}{4}$         &0.6096           &80 \& 256                         &0.7368\\
\hline
$\frac{1}{3}$         &0.6532           &80 \& 512                         &0.7425\\
\hline
$\mathbf{\frac{1}{2}}$&$\mathbf{0.7068}$&100 \& 128                        &0.6870\\
\hline
$\frac{2}{3}$         &0.6341           &$\mathbf{100}$ \& $\mathbf{256}$  &$\mathbf{0.7510}$\\
\hline
$\frac{3}{4}$         &0.6539           &100 \& 512                        &0.7438\\
\hline
\end{tabular}
\begin{tabular}[t]{|c|c|}
\hline
$\textbf{\#}$          &$ \textbf{F1}$    \\
$ \textbf{iter} $      &$ \textbf{Score}$ \\
\hline
0            &0.7510           \\
\hline
1            &0.7742           \\
\hline
3            &0.7745           \\
\hline
$\mathbf{5}$ &$\mathbf{0.7932}$\\
\hline
\end{tabular}
\label{table_params}
\end{table}
\vspace{-1em}
\subsection{Parameter Optimization}
The parameters of the proposed architecture are optimized only by training several IHFs based on fixed size parts. These experiments are performed on the \enquote{Coffee Cup} dataset.
\subsubsection{The size of the parts extracted during the initial registration}
The initial registration hypothesis is used by the \textit{iterative pose refinement} in order to improve the alignment's confidence (see Eq. \ref{Eq13}), and hence, $g^0$ is one of the important parameters that directly affects the success of the complete architecture. IHF-fixed size uses the parts that are of the same size during both the \textit{initial registration} and the \textit{iterative pose refinement}. IHF-variable size roughly aligns the object of interest during the \textit{initial registration} extracting coarsest parts, the ones occupying the largest area in image pixels. It iteratively refines this alignment based on the automatically extracted finer (smaller) parts, that is, it works in a size range. Thus, evaluating the performances of the \textit{initial registrations} for different part sizes is a crucial experiment that determines not only the optimum $g^0$ values, but also the range of the part sizes at which the IHF-variable size works in the most feasible way. The effect of the part size is examined by setting the IBS resolution $N$ to $80$, the HoCP feature dimension $d$ to $128$ in addition to the previously defined forest parameters. We change the part size $g^0$: $0.20$, $0.25$, $0.33$, $0.50$, $0.66$ and $0.75$ times of the object bounding box and for each, we train separate IHF-fixed size. The resultant Precision-Recall (PR) curves of the \textit{initial registrations} are shown in Fig. \ref{fig5} (a) and the corresponding F1 scores are demonstrated in Table \ref{table_params}. According to this figure and their corresponding F1 scores, we can choose any part size apart from the ones smaller than $\frac{1}{5}$ times of the bounding box. Considering both the computational load and the accuracy, we choose $\frac{1}{2}$ as the optimal part size for the IHF-fixed size. On the other hand, IHF-variable size uses the parts at various sizes, beginning with the coarsest (largest) ones extracted during the \textit{initial registration}, and ending with the finest (smallest) ones extracted at the last iteration of the \textit{iterative pose refinement}. We reanalyse the F1 scores presented in Table \ref{table_params} taking into account this size variation. When we increase the part size from $\frac{1}{5}$ to $\frac{3}{4}$, the F1 score ranges between $0.6$ and $0.7$. Despite the significant variation of the part size, the deviation in the F1 scores is negligible. We choose $\frac{3}{4}$ as the size of the parts extracted during the \textit{initial registration} phase of the IHF-variable size. One might suggest us to train both IHF-fixed and IHF-variable size separately in order to find the best corresponding $g^0$ values, however, it does not make sense. Because, the positive impact of the training on variable size parts is much observed during the iterative pose refinement phase. Hence, training only on the fixed size parts and the examination through the resultant F1 scores are reasonable to infer the best $g^0$ for each approach.
\begin{figure*}[!t]
\captionsetup[subfigure]{labelformat=empty}
\centering
\includegraphics[height=4.6in]{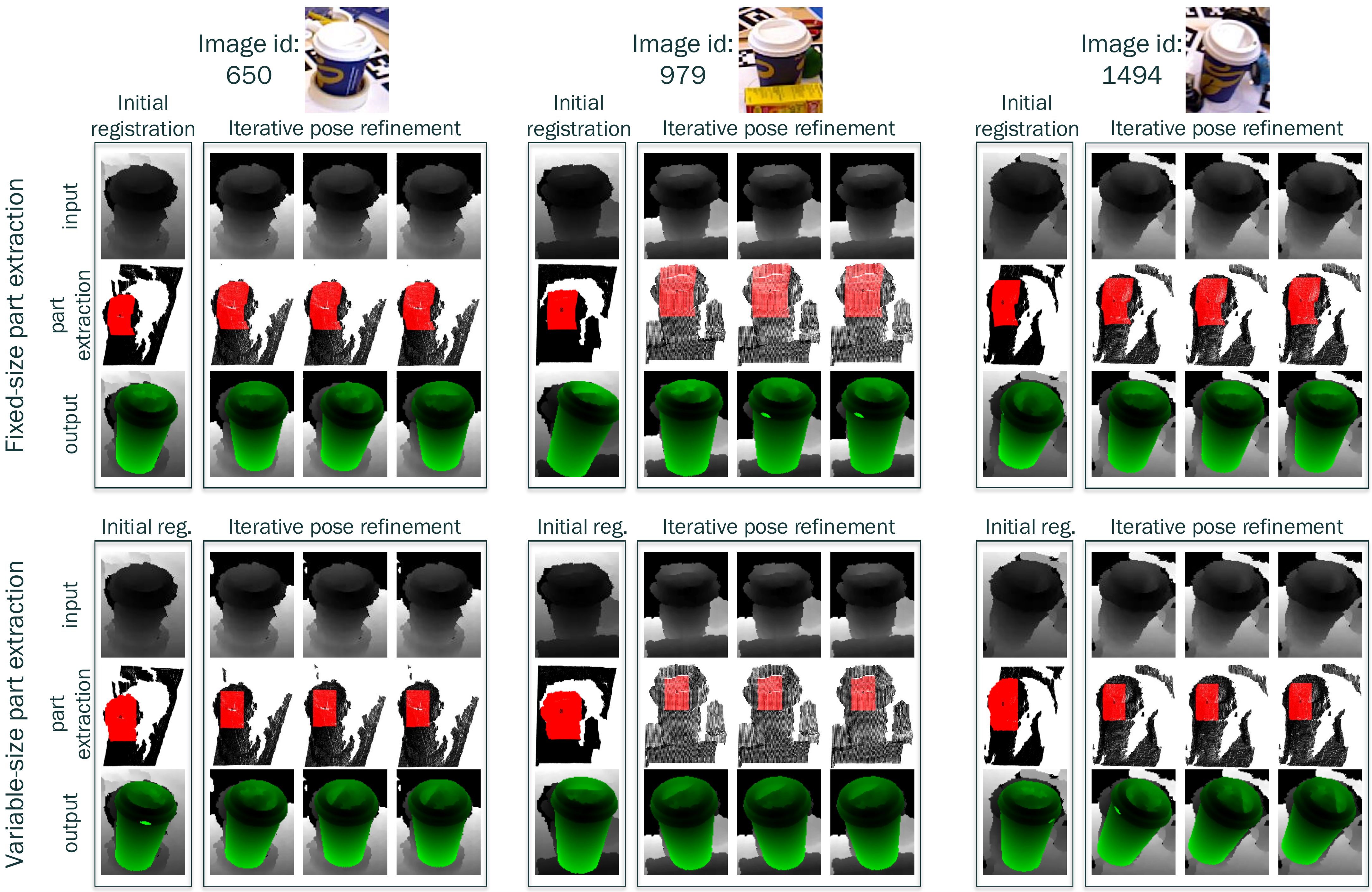}
\vspace*{-4mm}
\caption{Comparison between the variable-size and the fixed-size part extraction processes and their effect on the registration results. \textit{Iterative pose refinement} modules illustrate the $1^{st}$, $3^{rd}$ and $5^{th}$ iteration from left to right (RGB correspondences of the test objects are for better visualization).}
\label{fig6}
\end{figure*}
\subsubsection{IBS resolution and HoCP feature dimension}
We next tune the IBS resolution $N$ and the HoCP feature dimension $d$ by setting the part size to $\frac{1}{2}$ . We test the combinations of $N = 80, 100$ and $d = 128, 256, 512$, the ones that are the most applicable $N-d$ pairs to represent $\frac{1}{2}$ patch size. The PR curves of these combinations are depicted in Fig. \ref{fig5} (b) and the corresponding F1 scores are illustrated in Table \ref{table_params}. According to these results, we infer that the combinations composed by $d = 128$ relatively underperform whilst the remaining have approximately the same F1 scores. We take into account both the memory consumption and the accuracy, and agree on the values of $N = 100$ \& $d = 256$.
\subsubsection{The effect of the iteration}
The last parameter we optimize is the iteration number. We test several IHFs-fixed size \cite{14} each of which has $k = 0, 1, 3,$ and $5$ iterations, respectively. Their PR curves are shown in Fig. \ref{fig5} (c). As expected, the forests that use greater number of iterations show better performances (see Table \ref{table_params}) since more discriminative features are extracted due to the noise removal process.\\
\indent Figure \ref{fig6} demonstrates the registration results of several test objects comparing the IHFs that are trained on both the fixed-size and the variable-size parts. The RGB correspondences of the test objects are shown at the top, and each \enquote{iterative pose refinement} module illustrates the $1^{st}$, $3^{rd}$ and $5^{th}$ iteration at its $1^{st}$, $2^{nd}$ and the $3^{rd}$ columns, respectively. The part samples shown in the \enquote{part extraction} rows are grown around the same data point. We first discuss the \enquote{image id: 650}. The object is initially aligned by extracting the parts that are of size $\frac{1}{2}$ for the fixed-size and $\frac{3}{4}$ for the variable-size approach. By using the initial registration output, background clutter is removed from the test image. The amount of the reduction is approximately the same for both approach. After reduction, the test image is updated and is assigned as the input for the $1^{st}$ iteration. IHF-fixed size keeps on extracting the parts that are of size $\frac{1}{2}$ till the last iteration, whilst the IHF-variable size grows finer (smaller) regions in proportion to the removed foreground/background clutter. One can infer that the variable size approach registers the object of interest slightly better than the IHF-fixed size. For the \enquote{image id: 979} and the \enquote{image id: 1494}, the same regions of the test images are removed as the iteration progresses. However, the IHF-variable size demonstrate better results for both objects. This comparison also verifies that we have selected the optimum $g^0$ value for each approach. As the iteration progresses, we observe smooth transitions between the estimated translation and rotation parameters.
\begin{figure*}[!t]
\captionsetup[subfigure]{labelformat=empty}
\centering
\includegraphics[height=2.9in]{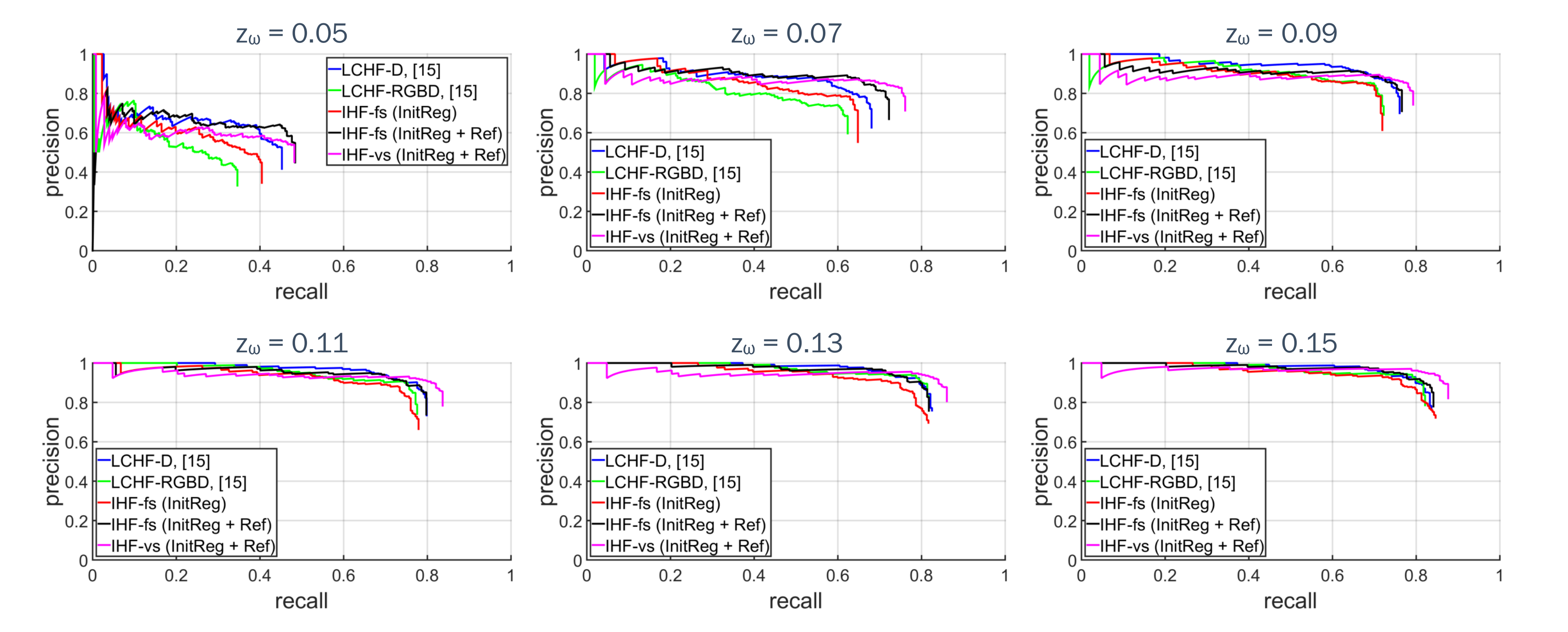}
\vspace*{-8mm}
\caption{Precision-Recall curves of the 'Coffee Cup' dataset: Each image compares the IHF-fixed size (initial registration), the IHF-fixed size (initial registration + iterative pose refinement) and the IHF-variable size (initial registration + iterative pose refinement) with the LCHFs \cite{15} trained separately on Depth, and on RGB-D channels. Greater values of $z_\omega$ result higher precision and recall values. F1 scores are presented in Table \ref{table_interclass_cup}.}
\label{fig7}
\end{figure*}
\begin{table*}[!t]
\caption{F1 scores of the 'Coffee Cup' dataset are shown at different $z_\omega$ values.}
\centering
\begin{tabular}{ |c |c| c| c| c| c| }
\hline
$\mathbf{z_\omega}$ & $\textbf{LCHF-D}$    & $\textbf{LCHF-RGBD}$ & $\textbf{IHF-fixed size (fs)}$ & $\textbf{IHF-fixed size (fs)}$ & $\textbf{IHF-variable size (vs)}$\\
$\textbf{value}$    & $\textbf{\cite{15}}$ & $\textbf{\cite{15}}$ & $\textbf{(InitReg)}$     & $\textbf{(InitReg + Ref)}$  & $\textbf{(InitReg + Ref)}$ \\
\hline
\multicolumn{1}{|c|}{}&
\multicolumn{5}{|c|}{F1 scores}\\
\hline
\multicolumn{1}{|c|}{0.05} & 0.4867 & 0.3818 & 0.4375 & $\mathbf{0.5297}$ & 0.5095 \\
\hline
\multicolumn{1}{|c|}{0.07} & 0.7202 & 0.6639 & 0.6985 & 0.7595 & $\mathbf{0.7891}$ \\
\hline
\multicolumn{1}{|c|}{0.09} & 0.7984 & 0.7683 & 0.7633 & 0.7975 & $\mathbf{0.8150}$ \\
\hline
\multicolumn{1}{|c|}{0.11} & 0.8344 & 0.8199 & 0.8000 & 0.8312 & $\mathbf{0.8565}$ \\
\hline
\multicolumn{1}{|c|}{0.13} & 0.8548 & 0.8554 & 0.8163 & 0.8481 & $\mathbf{0.8773}$ \\
\hline
\multicolumn{1}{|c|}{0.15} & 0.8589 & 0.8595 & 0.8353 & 0.8608 & $\mathbf{0.8940}$ \\
\hline
\end{tabular}
\label{table_interclass_cup} 
\end{table*}
\vspace{-1em}
\subsection{Comparative Study}
These experiments are conducted on the \enquote{Coffee Cup}, \enquote{Camera} and \enquote{Shampoo} datasets to compare our approach with the state-of-the-art methods including the Latent-Class Hough forests (LCHF) \cite{5} trained separately on the surface normal (LCHF-Depth (D) channel) and the color gradient + the surface normal (LCHF-RGBD channel) features. In order to make a fair comparison between methods, we train and test the LCHFs by using the authors' software. The forest parameters of all approaches are the same. For both datasets we generate PR curves at various $z_\omega$ values, beginning with the value that strictly limits the deviations between the ground truth and the estimated pose parameters, $0.05$, going up in $0.01$ increments, and ending with the value that accepts relatively rough estimations as correct, $0.15$.\\
\begin{figure*}[!t]
\captionsetup[subfigure]{labelformat=empty}
\centering
\includegraphics[height=2.7in]{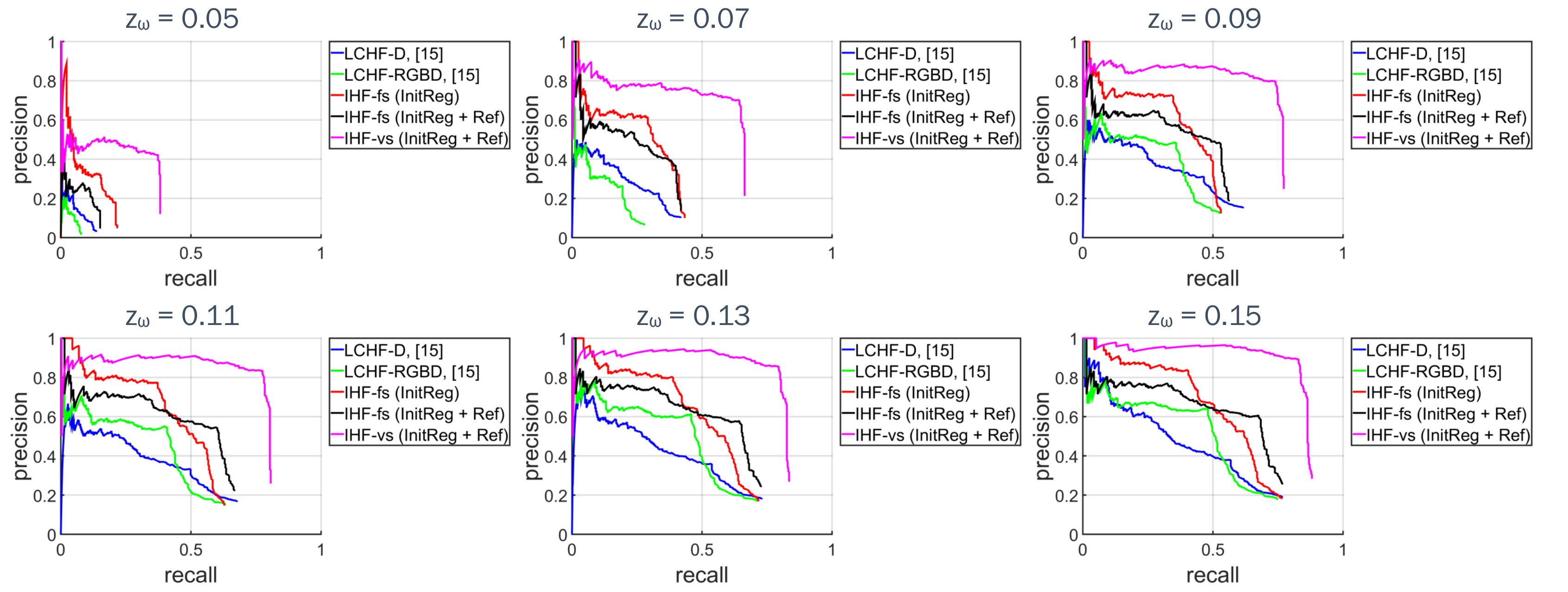}
\vspace*{-8mm}
\caption{Precision-Recall curves of the 'Camera' dataset: Each image compares the IHF-fixed size (initial registration), the IHF-fixed size (initial registration + iterative pose refinement) and the IHF-variable size (initial registration + iterative pose refinement) with the LCHFs \cite{15} trained separately on Depth, and on RGB-D channels. Greater values of $z_\omega$ result higher precision and recall values. F1 scores are presented in Table \ref{table_interclass_camera}.}
\label{fig8}
\end{figure*}
\begin{table*}[!t]
\caption{F1 scores of the 'Camera' dataset are shown at different $z_\omega$ values.}
\centering
\begin{tabular}{ |c |c| c| c| c| c| }
\hline
$\mathbf{z_\omega}$ & $\textbf{LCHF-D}$ & $\textbf{LCHF-RGBD}$ & $\textbf{IHF-fixed size (fs)}$ & $\textbf{IHF-fixed size (fs)}$ & $\textbf{IHF-variable size (vs)}$ \\
$\textbf{value}$    & $\textbf{\cite{15}}$     & $\textbf{\cite{15}}$     & $\textbf{(InitReg)}$ & $\textbf{(InitReg + Ref)}$  & $\textbf{(InitReg + Ref)}$ \\
\hline
\multicolumn{1}{|c|}{}&
\multicolumn{5}{|c|}{F1 scores}\\
\hline
\multicolumn{1}{|c|}{0.05} & 0.1003 & 0.0736 & 0.2071 & 0.1538 & $\mathbf{0.3963}$ \\
\hline
\multicolumn{1}{|c|}{0.07} & 0.2696 & 0.2240 & 0.3954 & 0.3878 & $\mathbf{0.6706}$ \\
\hline
\multicolumn{1}{|c|}{0.09} & 0.3723 & 0.4121 & 0.4761 & 0.5047 & $\mathbf{0.7680}$ \\
\hline
\multicolumn{1}{|c|}{0.11} & 0.3991 & 0.4674 & 0.5140 & 0.5710 & $\mathbf{0.8035}$ \\
\hline
\multicolumn{1}{|c|}{0.13} & 0.4304 & 0.5246 & 0.5494 & 0.6091 & $\mathbf{0.8242}$ \\
\hline
\multicolumn{1}{|c|}{0.15} & 0.4551 & 0.5500 & 0.5731 & 0.6388 & $\mathbf{0.8596}$ \\
\hline
\end{tabular}
\label{table_interclass_camera} 
\end{table*}
\begin{figure*}[!t]
\captionsetup[subfigure]{labelformat=empty}
\centering
\includegraphics[height=2.7in]{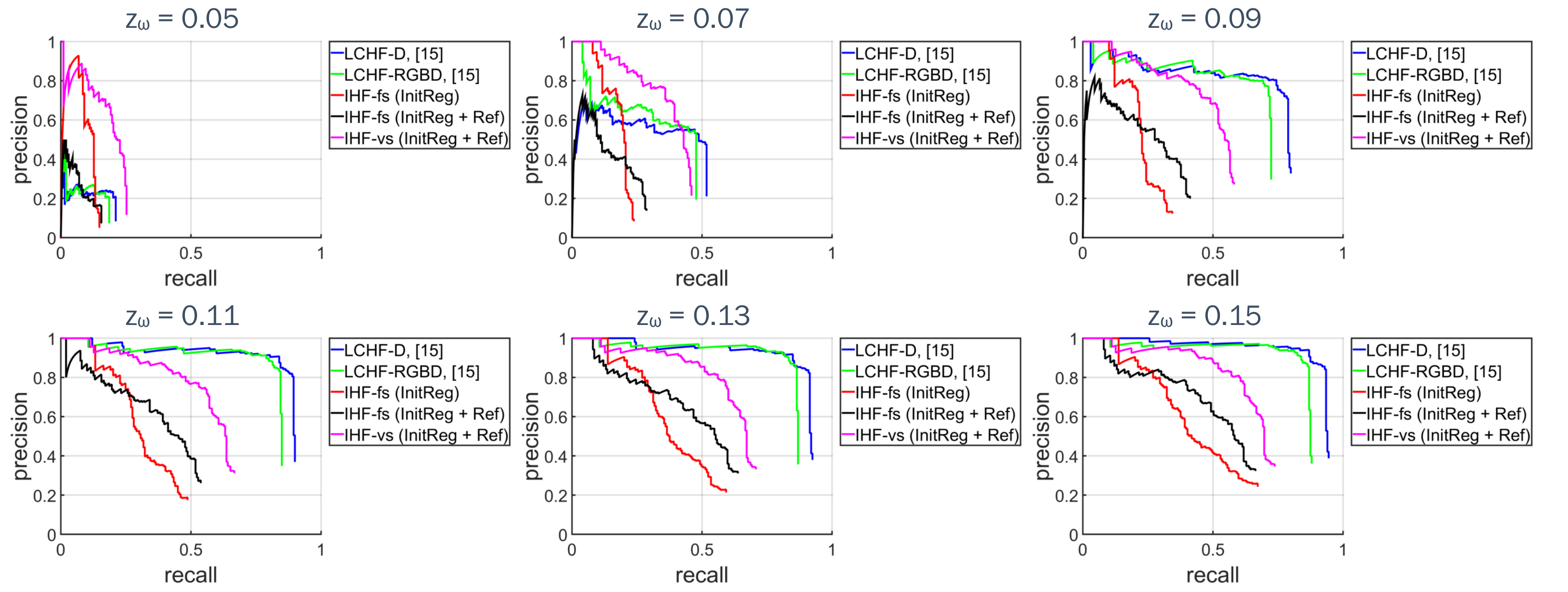}
\vspace*{-8mm}
\caption{Precision-Recall curves of the 'Shampoo' dataset: Each image compares the IHF-fixed size (initial registration), the IHF-fixed size (initial registration + iterative pose refinement) and the IHF-variable size (initial registration + iterative pose refinement) with the LCHFs \cite{15} trained separately on Depth, and on RGB-D channels. Greater values of $z_\omega$ result higher precision and recall values. F1 scores are presented in Table \ref{table_interclass_shampoo}.}
\label{fig8}
\end{figure*}
\begin{table*}[!t]
\caption{F1 scores of the 'Shampoo' dataset are shown at different $z_\omega$ values.}
\centering
\begin{tabular}{ |c |c| c| c| c| c| }
\hline
$\mathbf{z_\omega}$ & $\textbf{LCHF-D}$ & $\textbf{LCHF-RGBD}$ & $\textbf{IHF-fixed size (fs)}$ & $\textbf{IHF-fixed size (fs)}$ & $\textbf{IHF-variable size (vs)}$ \\
$\textbf{value}$    & $\textbf{\cite{15}}$     & $\textbf{\cite{15}}$     & $\textbf{(InitReg)}$ & $\textbf{(InitReg + Ref)}$  & $\textbf{(InitReg + Ref)}$ \\
\hline
\multicolumn{1}{|c|}{}&
\multicolumn{5}{|c|}{F1 scores}\\
\hline
\multicolumn{1}{|c|}{0.05} & 0.2168 & 0.197 & 0.2051 & 0.1597 & $\mathbf{0.3125}$ \\
\hline 
\multicolumn{1}{|c|}{0.07} & $\mathbf{0.5094}$ & 0.5067 & 0.2983 & 0.2811 & 0.50 \\
\hline
\multicolumn{1}{|c|}{0.09} & $\mathbf{0.7728}$ & 0.7439 & 0.3306 & 0.3819 & 0.5862 \\
\hline
\multicolumn{1}{|c|}{0.11} & $\mathbf{0.8720}$ & 0.8463 & 0.3878 & 0.4785 & 0.6379 \\
\hline
\multicolumn{1}{|c|}{0.13} & $\mathbf{0.8825}$ & 0.8670 & 0.4437 & 0.5436 & 0.6724 \\
\hline
\multicolumn{1}{|c|}{0.15} & $\mathbf{0.9033}$ & 0.8723 & 0.4713 & 0.5692 & 0.6898 \\
\hline
\end{tabular}
\label{table_interclass_shampoo} 
\end{table*}
\begin{table*}[!t]
\caption{F1 scores of the \enquote{Coffee Cup} and the \enquote{Camera} datasets are shown. These scores are the average of all $z_{\omega}$ that take each value in the range of [0.05 - 0.15].}
\centering
\begin{tabular}{ |c |c| c| c| c| c| }
\hline
$\textbf{Object}$ & $\textbf{LCHF-D}$ & $\textbf{LCHF-RGBD}$ & $\textbf{IHF-fixed size (fs)}$ & $\textbf{IHF-fixed size (fs)}$ & $\textbf{IHF-variable size (vs)}$ \\
                  & $\textbf{\cite{15}}$     & $\textbf{\cite{15}}$     & $\textbf{(InitReg)}$ & $\textbf{(InitReg + Ref)}$  & $\textbf{(InitReg + Ref)}$ \\
\hline
\multicolumn{1}{|c|}{}&
\multicolumn{5}{|c|}{F1 scores}\\
\hline
\multicolumn{1}{|c|}{$\textbf{Coffee Cup}$} & 0.7744 & 0.7410 & 0.7419 & 0.7834 & $\mathbf{0.8026}$ \\
\hline
\multicolumn{1}{|c|}{$\textbf{Camera}$} & 0.3441 & 0.3850 & 0.4631 & 0.4881 & $\mathbf{0.7323}$ \\
\hline
\multicolumn{1}{|c|}{$\textbf{Shampoo}$} & $\mathbf{0.7067}$ & 0.6870 & 0.3577 & 0.4067 & 0.5747 \\
\hline
\end{tabular}
\label{table_interclass_aver} 
\end{table*}
\indent The PR curves of the coffee cup dataset are depicted in Fig. \ref{fig7} for several $z_\omega$ values and their corresponding F1 scores are presented in Table \ref{table_interclass_cup}. A short analysis on the images of Fig. \ref{fig7} reveals that the increase in $z_\omega$ value generates higher F1 scores for each approach. According to Table \ref{table_interclass_cup}, the LCHF trained on the color gradient + surface normal features underperforms the LCHF trained merely on the surface normals. The main reason of this underperformance is the distortion along the object borders arising from the occlusion and the clutter, that is, the distortion of the color gradient information in the test process. The initial registration of the IHF-fixed size performs better than any versions of the LCHF. This shows that the HoCP features represent the parts better than the surface normals and color gradients providing robustness across occlusion, clutter and missing depth pixels. When this initial registration (see the $3^{th}$ column of Table \ref{table_interclass_cup}) acquired from the IHF-fixed size is iteratively refined, more accurate registrations are resulted (see the $4^{th}$ column of Table \ref{table_interclass_cup}). Because, the \textit{iterative pose refinement} module of the IHF-fixed size reduces the amount of the noise in the test depth maps removing foreground/background clutter. This removal process also enables IHF-fixed size to compute more discriminative control points for better shape representation. The IHF-variable size with initial registration + iterative pose refinement outperforms other approaches. The main reason of this high performance is the utilization of the parts that are different in size. The cascaded representation of the locality increases the robustness across clutter, occlusion, missing depth pixels and/or similar looking distractors. The object of interest is first roughly aligned by extracting the coarsest parts. It is highly possible that these initially extracted parts include the portions belonging to the background/foreground clutter since they are the coarsest and are close to a holistic template. Despite the fact that we apply a depth check in order to get rid of those noise during testing, it is highly naive. On the other hand, the proposed framework can get rid of those structural perturbations by growing smaller regions as the iteration progresses. Apart from that, the control point descriptors computed at later iterations allows the complete framework to represent the shapes in a more discriminative manner.\\
\indent Regarding the camera dataset, we show its PR curves in Fig. \ref{fig8} for several $z_\omega$ values and the corresponding F1 scores in Table \ref{table_interclass_camera}. The approaches under comparison perform worse on this dataset with respect to the coffee cup. Unlike the results obtained from the coffee cup dataset, we see the positive impact of the color gradients when they are utilized along with the surface normals at most of the $z_\omega$ values (compare the $1^{st}$ and the $2^{nd}$ columns of Table \ref{table_interclass_camera}). IHF-fixed size register objects more accurate than any versions of the LCHF thanks to the utilization of the discriminative information embedded into the scale-variant HoCP features and the iterative refinement of the test depth maps. IHF-variable size significantly outperforms other approaches demonstrating the importance of the simultaneous utilization of variable size parts. The HoCP representations of the cascaded regions grown around the same data points allow the algorithm to be aware of occlusion, clutter and missing depth values. More confident registrations are hypothesized as the iteration progresses based on more discriminative representations of the smaller parts. The registration performance of the proposed architecture is shown in Figure \ref{fig8} for the shampoo object and the corresponding F1 scores are demonstrated in Table \ref{table_interclass_shampoo} for varying $z_\omega$ values. In cases of registering at lower $z_\omega$ values, our approach shows better performance than the LCHFs, however, when we accept relatively rough estimations as correct, \textit{i.e.}, higher $z_\omega$s, our approach underperforms.\\
\indent Since we address the registration problem rather than individual object detection or pose estimation, we integrate the effect of the different error ratios into our comparisons. We average the F1 scores that are computed at each $z_\omega$ in the range of [0.05-0.15] and report in Table \ref{table_interclass_aver}. Figure \ref{fig9} illustrates several accurate registrations hypothesized by the proposed framework on the camera and the coffee cup objects. We further evaluate the performance of the globally optimized ICP algorithm proposed in \cite{21} on our test dataset. We use the software kindly provided by the authors and set the  default parameters. While accurate registration results are obtained on the clean dataset, it diverges in the case of highly occluded and cluttered point clouds (see Fig. \ref{fig10}).
\begin{figure*}[!t]
\captionsetup[subfigure]{labelformat=empty}
\centering
\includegraphics[height=8.9in]{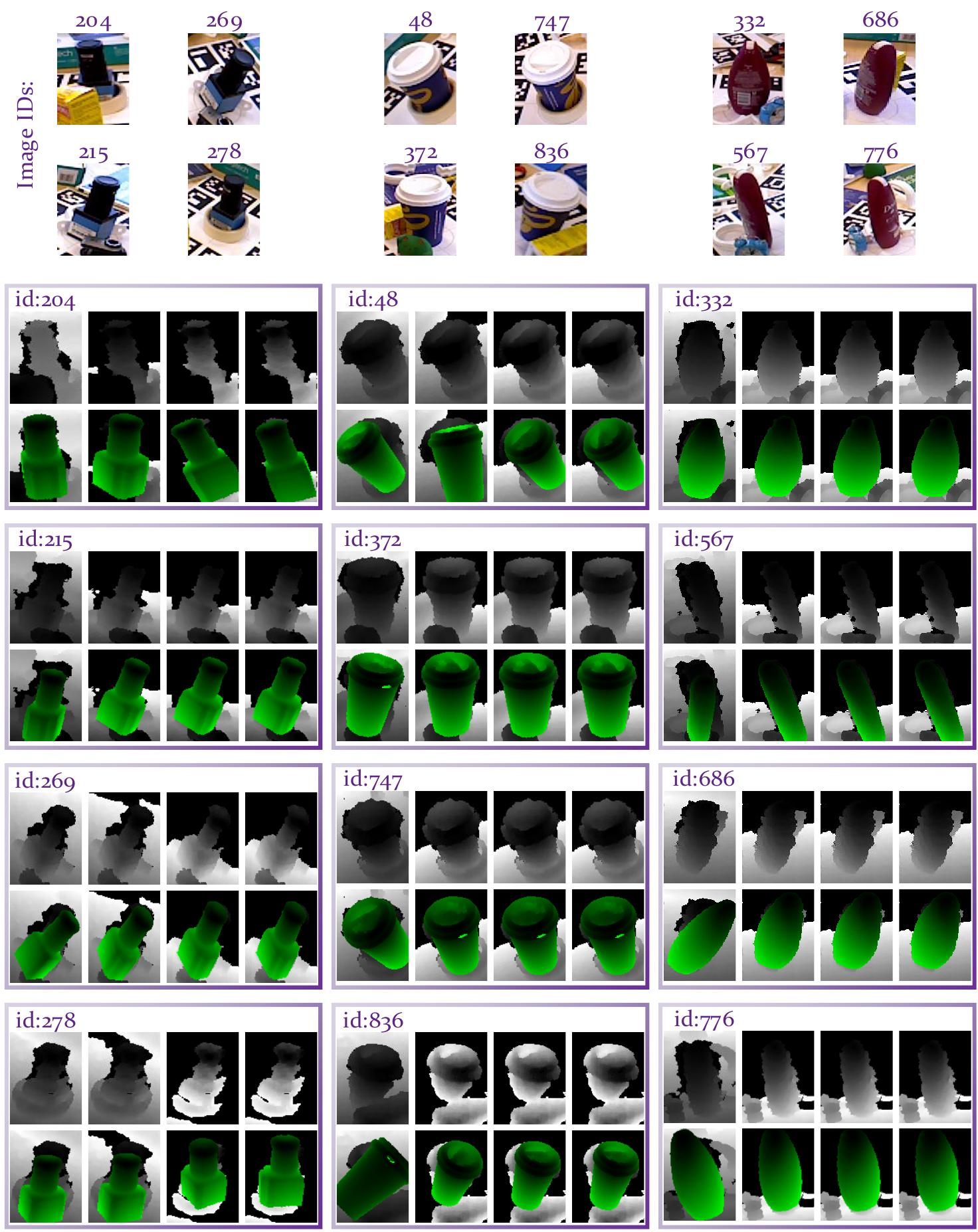}
\vspace*{-2mm}
\caption{Some qualitative results. For each octonary: the $1^{st}$ column illustrates the test image and the initial hypothesis (\textit{initial registration}) and the remaining columns demonstrate the $1^{st}$, the $3^{rd}$ and the $5^{th}$ iterations (\textit{iterative pose refinement}). The test images are updated by removing the background/foreground clutter.}
\label{fig9}
\end{figure*}
\section{Discussion and Conclusion}
\label{conclusion}
In this study, we have proposed a novel architecture, \textit{Iterative Hough Forest with Histogram of Control Points}, addressing 6D object registration rather than individually estimating either the object's location in a 2D/3D bounding box, or the object's orientation (roll, pitch, yaw). Any off-the-shelf detector can accurately provide a coarse 2D or 3D bounding box for the object of interest. Various object orientation predictors are also available, however, they depend on clearly segmented target objects. Our architecture fundamentally targets to eliminate the shortcomings of these individual detectors and orientation predictors estimating occluded and cluttered objects' 6D pose given a candidate 2D bounding box. Our IHF is learnt using parts extracted only from the positive samples. These parts are represented with scale-variant HoCP features, which we derive from recently introduced Implicit B-Splines (IBS).\\
\indent At test time, we apply two different strategies regarding the parts used to train the forest: The first strategy we apply roughly aligns the object  and iteratively refines this alignment based on more discriminative control point descriptors that are computed due to the elimination of background/foreground clutter. The part size is fixed and is empirically predefined. On the other hand, the predefined part size might not be generalizable enough across different objects, degrading the registration performance of the proposed study on one object while working well on another one. Besides, discriminative information encoded into small sized parts might not be fully exploited by larger parts, most particularly when the object representation is scale-variant. Inspiring by these observations, we use variable size parts in the second strategy. An automatic variable size part extraction framework iteratively refines the object's initial pose that is roughly aligned due to the extraction of coarsest parts, the ones occupying the largest area in image pixels. The iterative refinement is accomplished based on finer (smaller) parts that are represented with more discriminative control point descriptors by using our \textit{Iterative Hough Forest}. The experimental results report that our approach show better registration performance than the state-of-the-art methods.
\begin{figure}[!t]
\captionsetup[subfigure]{labelformat=empty}
\centering
\subfloat{\includegraphics[width=3.2in]{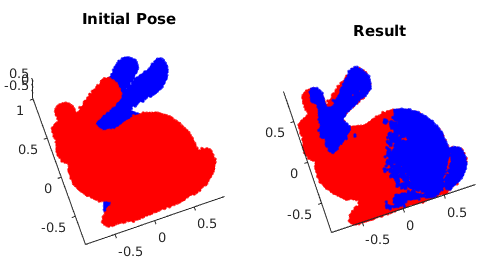}}

\subfloat{\includegraphics[width=3.2in]{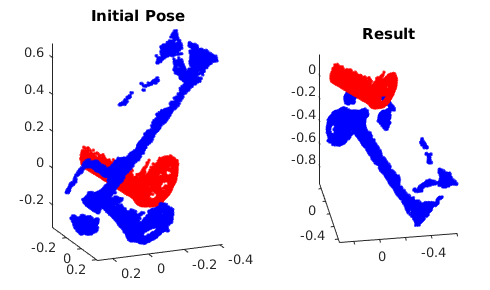}}
\caption{The performance of the Go-ICP \cite{21} algorithm on our dataset: Despite it achieves good registration on the clean point cloud, it diverges on our dataset.}
\label{fig10}
\end{figure}
\section{Acknowledgements}
\label{Acknowledgements}
Caner Sahin is funded by the Turkish Ministry of National Education.

\end{document}